\documentclass{article}
\usepackage{ijcai25}

\usepackage{times}
\usepackage{soul}
\usepackage{url}
\usepackage[hidelinks]{hyperref}
\usepackage[utf8]{inputenc}
\usepackage[small]{caption}
\usepackage{graphicx}
\usepackage{amsmath}
\usepackage{amsthm}
\usepackage{booktabs}
\usepackage{algorithm}
\usepackage{algorithmic}
\usepackage[switch]{lineno}

\usepackage{xcolor}
\usepackage{amsmath}
\usepackage{amssymb}
\usepackage{mathtools}
\usepackage{amsthm}
\usepackage{multirow}

\title{Enhancing Multimodal Protein Function Prediction Through Dual-Branch Dynamic Selection with Reconstructive Pre-Training}

\author{
    Anonymous Submission
    \affiliations
    Paper 1116
    \emails
}

\author{
Xiaoling Luo$^1$\and
Peng Chen$^2$\and
Chengliang Liu$^3$\and
Xiaopeng Jin$^{4}$\thanks{Co-corresponding authors: Xiaopeng Jin, Jie Wen.}\and
Jie Wen$^{5*}$\and
Yumeng Liu$^4$\And
Junsong Wang$^4$\\
\affiliations
$^1$College of Computer Science and Software Engineering, Shenzhen University, Shenzhen, China\\
$^2$College of Applied Technology, Shenzhen University, Shenzhen, China\\
$^3$Laboratory for Artificial Intelligence in Design, Hong Kong\\
$^4$College of Big Data and Internet, Shenzhen Technology University, Shenzhen, China\\
$^5$College of Computer Science and Technology, Harbin Institute of Technology, Shenzhen, China\\
\emails
xiaolingluoo@outlook.com,
2300411008@emal.szu.edu.cn,
liucl1996@163.com,
jinxiaopengit@gmail.com,
wenjie@hit.edu.cn,
liuyumeng@sztu.edu.cn,
wangjunsong@sztu.edu.cn
}

\begin{document}

\maketitle

\begin{abstract}
   
Multimodal protein features play a crucial role in protein function prediction. However, these features encompass a wide range of information, ranging from structural data and sequence features to protein attributes and interaction networks, making it challenging to decipher their complex interconnections. In this work, we propose a multimodal protein function prediction method (DSRPGO) by utilizing dynamic selection and reconstructive pre-training mechanisms. To acquire complex protein information, we introduce reconstructive pre-training to mine more fine-grained information with low semantic levels. Moreover, we put forward the Bidirectional Interaction Module (BInM) to facilitate interactive learning among multimodal features. Additionally, to address the difficulty of hierarchical multi-label classification in this task, a Dynamic Selection Module (DSM) is designed to select the feature representation that is most conducive to current protein function prediction. Our proposed DSRPGO model improves significantly in BPO, MFO, and CCO on human datasets, thereby outperforming other benchmark models.

\end{abstract}

\section{Introduction}
\label{submission}

Protein function prediction has become a key challenge in biology, with the rapid development of bioinformatics \cite{hasselgren2024artificial}. The Gene Ontology (GO) framework \cite{MA2025125366} standardizes protein functions into three categories: biological process (BPO), molecular function (MFO), and cellular component (CCO) \cite{gene2023gene}. In recent decades, numerous deep learning methods \cite{you2021deepgraphgo,zhang2023hnetgo} have been developed to predict protein functions. However, using single-modal features often faces data limitations \cite{kulmanov2020deepgoplus}.
Many studies \cite{fan2020graph2go} have shown that using protein sequence information significantly improves the accuracy of MFO. Still, many proteins share functional similarities but have dissimilar sequences \cite{bhzs}. As a result, for proteins with low sequence similarity, the accuracy of predictions may be compromised. Moreover, structure-based methods usually perform better, but the high complexity of protein structures and data acquisition costs limit their application \cite{paysan2023interpro}. Furthermore, the noise introduced during the generation of protein-protein interaction (PPI) networks \cite{9387128} through high-throughput techniques poses risks to the accuracy of predictions \cite{chen2024dualnetgo}.

Therefore, integrating these different types of protein data and taking advantage of their complementary advantages in functional prediction is an important way \cite{10822064} to improve the performance of protein function prediction. These methods mainly adopt two strategies: graph neural networks (GNNs) \cite{you2021deepgraphgo} and autoencoders \cite{gligorijevic2018deepnf,fan2020graph2go,pan2023pfresgo}.
Graph2GO \cite{fan2020graph2go} integrates sequence similarity and PPI networks using GNNs, treating protein sequences and structures as node features. However, those using GNNs \cite{8983075} may amplify noise and face issues with over-smoothing. To address these limitations, CFAGO \cite{wu2023cfago} introduces Transformer-based fusion within autoencoders to enhance multimodal feature integration.

However, current multimodal approaches mainly fuse information without exploring the potential complementarity between different modalities. To address this issue, we propose a multimodal method for protein function prediction that efficiently mines the complex internal relationships among spatial structure features, such as PPI networks, subcellular locations, and protein domains, as well as sequence features, specifically the amino acid sequence. 
Furthermore, due to the complexity of protein information, existing models tend to ignore the detailed features inside the information, such as PPI local network topology, connection strength, amino acid frequency distribution, and key sequence fragments. We add a reconstruction pre-training step to obtain more low-semantic and fine-grained features from protein information of multiple modes. By learning these basic features, the model provides a richer representational basis for downstream tasks.

In addition, large language models play an important role in improving protein function prediction.
Inspired by large language models, the protein sequence information in our method is extracted using the pre-trained ProtT5 \cite{prott5}. In this work, to better learn multimodal information, our proposed DSRPGO model includes a shared and an interactive learning branch. In the shared learning branch, we concatenate features from different modalities and perform joint analysis in a unified representation space. Moreover, we introduce the Bidirectional Interaction Module (BInM), where each modality both influences and receives information from others, enhancing overall understanding.

Besides, faced with thousands of protein functions, accurately predicting the protein function of a sample remains a challenging issue. Protein function prediction is essentially a complex hierarchical multi-label classification problem. In this situation, we propose the Dynamic Selection Module (DSM) to dynamically select the optimal feature combination for fitting more diverse protein functions. The code and supplementary materials have been open-sourced\footnote{\url{https://github.com/kioedru/DSRPGO}}. Our main contributions can be summarized as follows:

\begin{itemize} 
\item We propose a multimodal feature-based approach for protein function prediction that overcomes the limitations of single-modality methods, effectively representing protein functional characteristics.
\item A reconstructive pre-training phase is designed to make the model capable of learning more low-semantic fine-grained features to assist the model in understanding protein function.
\item Our proposed BInM incorporates a bidirectional interaction mechanism to promote efficient fusion and information exchange between sequence and spatial features, enhancing the model's ability to capture strong protein information between different modes.
\item We construct the DSM that enables the model to adaptively select channel features most relevant to specific functional labels, resulting in enhanced performance.
\end{itemize}

\section{Methodology}
\begin{figure*}[!t]
\centering
\includegraphics[width=16.5cm]{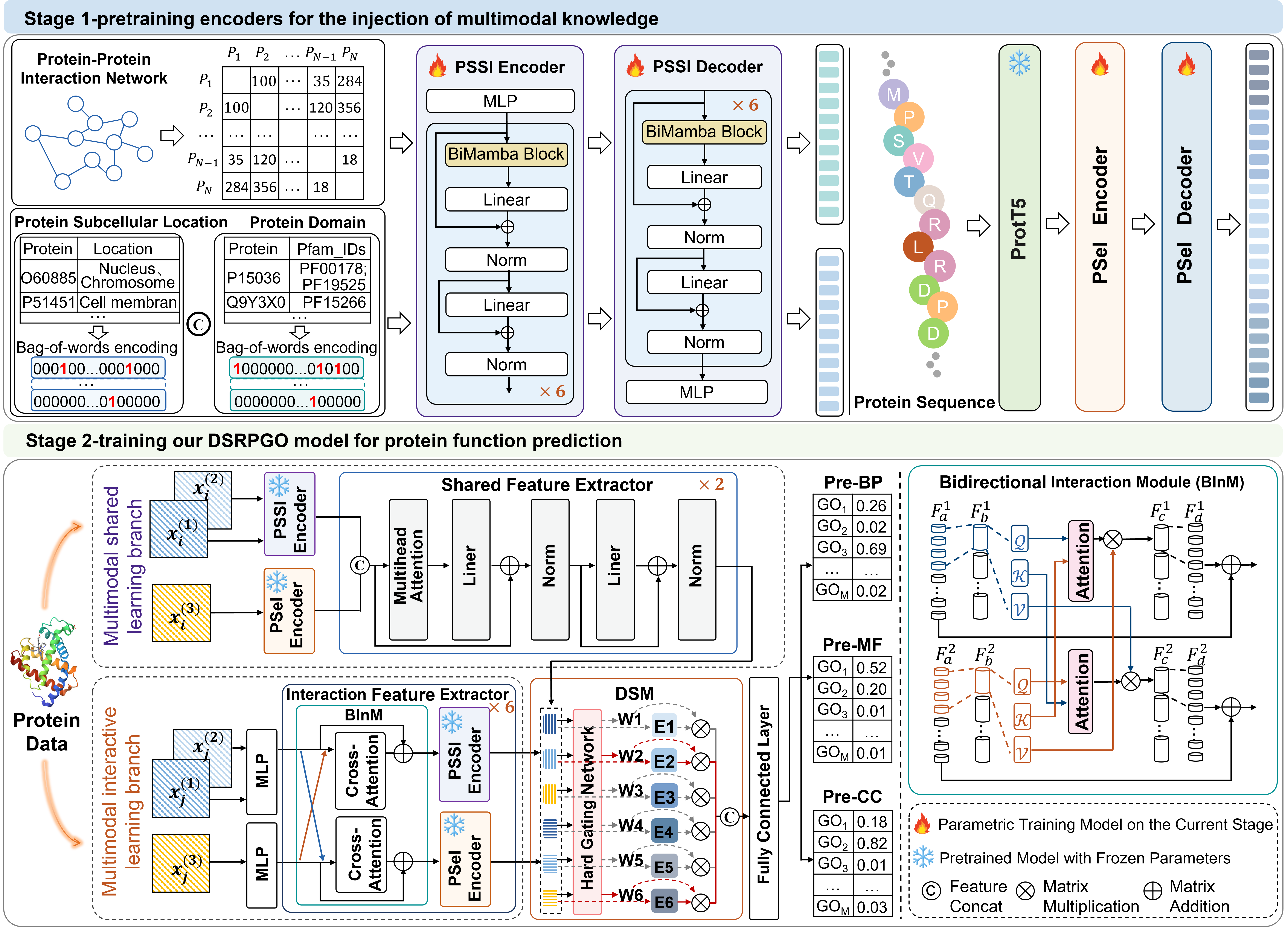}
\caption{An illustration of our proposed method. This method is mainly divided into two stages. The first stage is to pre-train the Protein Spatial Structure Information (PSSI) encoder and Protein Sequence Information (PSeI) encoder for the injection of multimodal knowledge
. The second stage is training our proposed DSRPGO model, which consists of an MSL-Branch, a MIL-Branch with the Bidirectional Interaction Module (BInM), and the Dynamic Selection Module (DSM).}
\label{main}
\end{figure*}

Our proposed method efficiently captures multimodal information about proteins through a strategy for two-step training. In the pre-training stage, we use the encoder-decoder model to learn and inject multimodal knowledge. For spatial features including PPI, subcellular location, and protein domains, a Protein Spatial Structured Information (PSSI) encoder-decoder model using the BiMamba blocks is introduced in this stage. To mine sequence features including protein sequences, we design a Protein Sequence Information (PSeI) encoder-decoder model based on the Transformer blocks for pre-training. Then, during our DSRPGO model training phase, we integrate and learn features from multimodal information. The proposed model is primarily divided into two major branches: one is the multimodal shared learning branch (MSL-Branch), and the other is the multimodal interactive learning branch (MIL-Branch). Protein data are processed through these branches to generate several sets of features, which serve as inputs for DSM. Finally, the model dynamic selects the optimal features for the current protein, to enhance performance in protein function prediction. An illustration of our proposed method can be seen in Figure \ref{main}. 

\subsection{Reconstructive Pre-training}
In the reconstructive pre-training stage, to obtain feature extractors that are good at mining fine-grained features from multi-modal protein information, we utilize the PSSI and PSeI encoder-decoder model for feature reconstruction.

\subsubsection{PSSI Encoder-Decoder Learning}
The PPI network gets an $N \times N$ adjacency matrix by matrix conversion as input to the encoder. Moreover, another input to the encoder is obtained by concatenating the bag-of-words encodings of subcellular location and Protein Domain.

\textbf{Mamba Preliminaries.}
Mamba \cite{gu2023mamba} extends the capabilities of the State-Space Models (SSMs) \cite{gu2022train} by enabling the transformation of a continuous 1D input \(x_t \in \mathbb{R}\) to \(y_t \in \mathbb{R}\) via a learnable hidden state \(h_t \in \mathbb{R}^{\hat{N}}\) with discrete parameters \(\bar{A} \in \mathbb{R}^{{\hat{N}} \times {\hat{N}}}\), \(\bar{B} \in \mathbb{R}^{1 \times {\hat{N}}}\), and \(\bar{C} \in \mathbb{R}^{1 \times {\hat{N}}}\) as follows:
{
\begin{equation}
\label{Eq1}
\begin{aligned}
    h_t &= \bar{A} h_{t-1} + \bar{B} x_t, ~y_t = C h_t + D h_t, \\
    \bar{A} &= e^{\Delta A}, ~\bar{B} = (\Delta A)^{-1} (e^{\Delta A} - I) \cdot \Delta B, ~\bar{C} = C.
\end{aligned}
\end{equation}
}

\(\bar{A} \) and \(\bar{B} \) are continuous \( A \) and \( B \) converted to discrete evolution parameters using a timescale parameter $\Delta$. To process discrete-time sequences sampled at intervals of $\Delta$, SSMs can be calculated using the recurrence formula. \(\bar{C} \) represents the projection parameters. In addition, the models compute output through a global convolution as follows:
{
\begin{equation}
\bar{K} = (\bar{C}\bar{B}, \bar{C}\bar{A}\bar{B}, \ldots, \bar{C}\bar{A}^{{\hat{N}}-1}\bar{B}), ~y = x * \bar{K},
\end{equation}
}
where $\hat{N}$ is the length of \(x\), and $\bar{K}$ is a convolutional kernel.

\begin{figure}[!t]
\centering
\includegraphics[width=8.5cm]{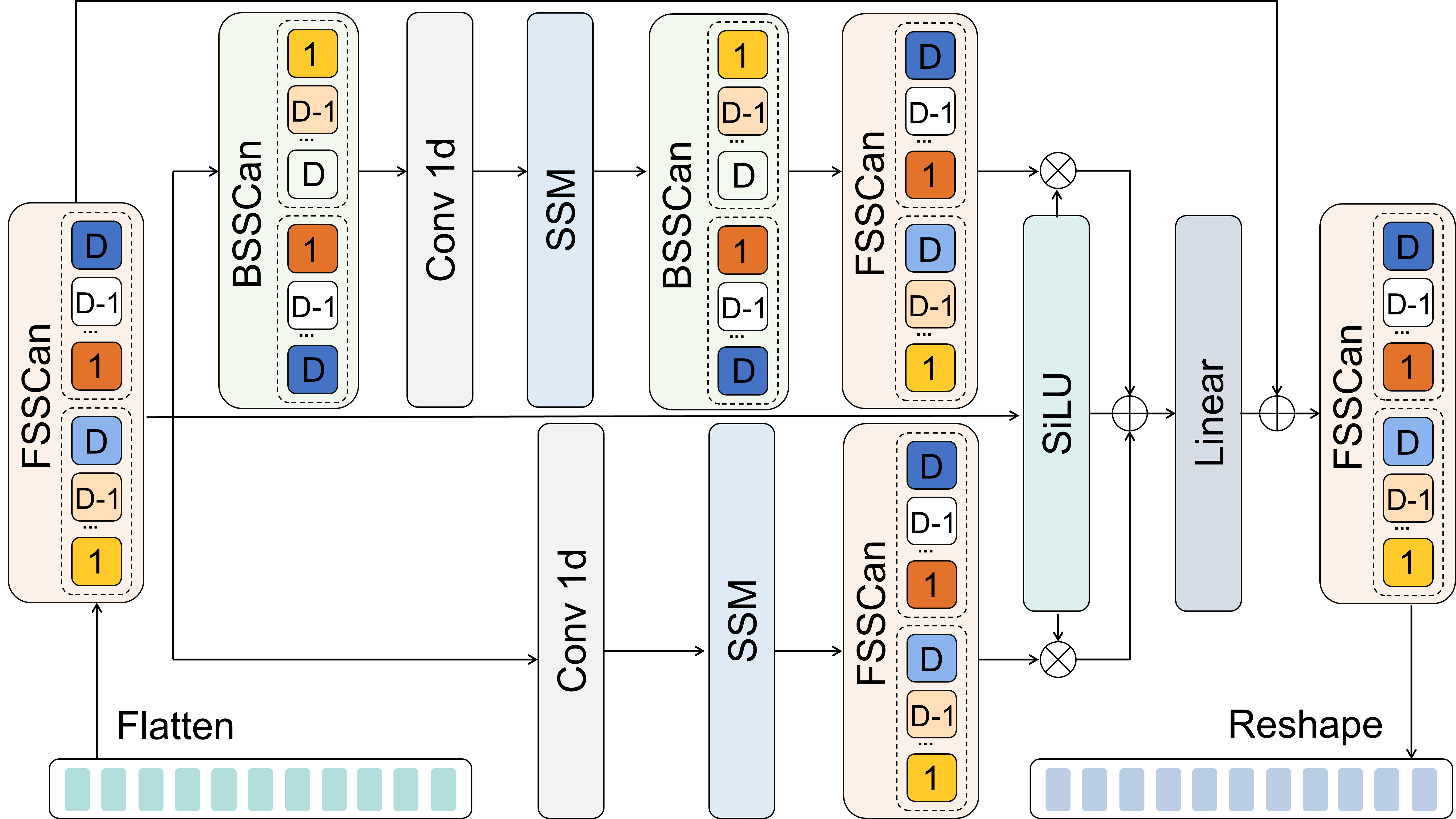}
\caption{Structure of the BiMamba block.}
\label{Mamba}
\end{figure}

\textbf{BiMamba Block. }
Inspired by the selective scan mechanism in Vision Mamba \cite{zhu2024vision}, BiMamba Block introduces a novel bidirectional selective scanning mechanism designed for protein data, capturing both the start and end of spatial structure features for enhanced detail and context. Multi-dimensional features are first converted into one-dimensional vectors. Features $x_{sp}$ from PPI, subcellular location, and protein domains are then passed through BiMamba blocks, interleaved with linear layers and residual operations. As shown in Figure \ref{Mamba}, forward (FSScan) and backward selective scans (BSScan) extract bidirectional matrix features via positional transformations and reconstructions. Transformed tokens are scanned using Equation \ref{Eq1} to produce new features, with BiMamba's output $\tilde{x}_{sp}$ expressed as:
{
\begin{align*}
\left\{
\begin{aligned}
\tilde{x}_{sp} = &  FSSCan(x_{sp})+FSSCan(Linear(F_{\alpha} \odot F_{\sigma}+F_{\beta}  \\
&  \odot F_{\sigma} +F_{\sigma} )), \\
F_{\alpha} = & \, FSSCan(BSSCan(SSM(Conv1d \\
& (BSSCan(FSSCan(x_{sp})))))), \\
F_{\beta} = & \, FSSCan(SSM(Conv1d(FSSCan(x_{sp})))), \\
F_{\sigma} = & \, SiLU(FSSCan(x_{sp})),
\end{aligned}
\right.
\end{align*}
}
where the operation $\odot$ denotes the Hadamard product.

\textbf{PSSI Encoder.}
In this section, we propose a PSSI encoder architecture designed to effectively map high-dimensional input data into a low-dimensional latent space. The PSSI encoder consists of multilayer perceptrons (MLPs), BiMamba block, Linear and Norm layers, which work in concert to extract features from the input data and generate a compact latent representation. Assume that the input feature $x_{i}^{h(k)} \in \mathrm{R}^{H_i^k}$ is a high-dimensional vector of the $i$-th protein, where $H_i^k$ represents the feature dimension of the $k$-th input source. This feature is reconstructed using  MLP to output a low-dimensional representation $x_{i}^{d(k)} \in \mathrm{R}^{D}$, where $D$ denotes the size of the MLP hidden layer.
\textbf{PSSI Decoder.}
The architecture of the PSSI decoder is a counterpart to that of the encoder. The PSSI decoder rebuilds the given protein spatial structure information based on the hidden representations output by the encoder. This process involves BiMamba computation and residual operations, optimizing the cross-entropy loss function to enhance the performance. 
After taking the output $x_{i}^{d(k)}$ of the PSSI encoder and passing through the BiMamba block, alternating Linear and Norm layers, we obtain the recovered high-dimensional features $\bar{x}_{i}^{h(k)} \in \mathrm{R}^{H_i^k}$.
The overarching objective of the encoder-decoder architecture is to minimize the sample wise binary cross-entropy loss between the original and reconstructed source features, thereby enhancing the model's predictive accuracy and fidelity in representing protein data. The loss function of PSSI encoder-decoder is:
{
\begin{align}
\mathcal{L}_{sp} = & \ \frac{1}{N} \sum_{i=1}^{N} \sum_{k=1}^{K} \sum_{j=1}^{H_i^k} 
    -\bigg[ x_{i j}^{h(k)} \log \bar{x}_{i j}^{h(k)} \nonumber \\
    & \ + \left(1 - x_{i j}^{h(k)}\right) \log \left(1 - \bar{x}_{i j}^{h(k)}\right) \bigg],
\end{align}
}
where $N$ is the number of total proteins, $K$ is the number of input sources, ${x}_{ij}^{h(k)}$ and $\bar{x}_{i j}^{h(k)}$ denotes the $j$-th dimension vector of $x_{i}^{h(k)}$ and $\bar{x}_{i}^{h(k)}$.
\subsubsection{PSeI Encoder-Decoder Learning}
In PSeI encoder-decoder, the transformer block with multi-head self-attention (MSA) mechanism \cite{Dosovitskiy2021} extracts long-distance features from protein sequences. Then, to further leverage these features, we use the pre-trained ProtT5 \cite{prott5} model to parse the protein sequences. 
To achieve this, we froze the parameters of ProtT5 and connected it to the PSeI encoder for further pretaining.

\textbf{PSeI Encoder.}
The PSeI encoder consists of an MLP block and $6$ self-attention blocks. The self-attention block includes an MSA computation layer, as well as alternating linear and norm layers, connected through a residual structure. Assuming the input of the self-attention block is $\tilde{s}_{i}^{d} = MLP(s_{i}^{h})$, the output feature is $\hat{s}_{i}^{d}\in \mathrm{R}^{D}$:
{
\begin{equation}
\resizebox{0.91\linewidth}{!}{$\hat{s}_{i}^{d}= N(N(\tilde{s}_{i}^{d}+L(MSA(\tilde{s}_{i}^{d})))+L(N(\tilde{s}_{i}^{d}+L(MSA(\tilde{s}_{i}^{d}))))),$}
\end{equation}
}
where $s_{i}^{h} \in \mathrm{R}^{H_i}$ is the $i$-th input sequence feature of encoder, and $H_i$ is the dimension of input feature.  $L(x)$ denotes the fuction of Linear layer, and $N(x)$ denotes the Norm layer.

\textbf{PSeI Decoder.}
The PSeI decoder takes the hidden states from the encoder as input, which contains compressed information about the input sequence. To obtain the final protein sequence encoding, we designed the PSeI decoder using a combination of $6$ self-attention blocks and one MLP block. Then, the output feature of the PSeI decoder is $\hat{s}_{i}^{h} \in \mathbb{R}^{H_i}$. Like the PSSI encoder-decoder, the loss function $\mathcal{L}_{se}$ for the PSeI encoder-decoder also adopts the form of cross-entropy:
{
\begin{equation}
\resizebox{0.91\linewidth}{!}{$\mathcal{L}_{se}=\frac{1}{N} \sum_{i=1}^{N} \sum_{j=1}^{H_i}-\left[s_{i j}^{h} \log \bar{s}_{i j}^{h}+\left(1-s_{i j}^{h}\right) \log \left(1-\bar{s}_{i j}^{h}\right)\right],$}
\end{equation}
}
where $i$ denotes the sequence input of the $i$-th protein, $j$ is the $j$-th dimension vector of the feature map.

\subsection{Bidirectional Interaction and Dynamic Selection for Protein Function Prediction}
In this section, we apply the encoders sensitive to low semantic features obtained in the pre-training stage to high semantic tasks. Specifically, to improve the performance of protein function prediction, BInM and DSM modules are proposed to capture deep interaction information between multimodal features and dynamically screen the features most suitable for the current task.
\subsubsection{Bidirectional Interaction Module} 
The proposed BInM enhances the model's ability to learn complex patterns by integrating information across modalities. Using cross-attention, it compares query (Q) vectors with key (K) vectors from the opposite branch, enabling bidirectional interaction. This approach captures interdependencies between branches more effectively, similar to MSA but focused on cross-branch connections.

Therefore, we assume that the features transformed by PPI are represented as $x_i^{(1)}$, and the features obtained from the encoding of subcellular location and protein domains are concatenated to form $x_i^{(2)}$, while the features extracted through the ProtT foundation model for protein sequences are denoted as $x_i^{(3)}$. Subsequently, $x_i^{(1)}$ and $x_i^{(2)}$ get features with the same dimension after the MLP reconstruction features, and their concatenated feature map $\widetilde{x}_i^B$ is used as the input of the first branch of BInM. Similarly, $\overline{x}_i^B$, the input to the second branch of BInM, is derived from $x_i^{(3)}$ after its transformation through the MLP. In BInM, the input embedded patches $F_a^{1}\in\mathbb{R}^{L_a \times D_a}$ and $F_a^{2}\in\mathbb{R}^{L_a \times D_a}$ are initially and randomly divided into multiple heads vectors $F_b^{1} \in\mathbb{R}^{L_a \times D_b \times H_b}$ and $F_b^{2} \in\mathbb{R}^{L_a \times D_b \times H_b}$, where $H_b$ is the number of multiple heads.

As shown in Figure \ref{main}, $F_b^{1}$ and $F_b^{2}$ are converted into queries $\mathcal{Q}^{1}(F_b^{1})$ and $\mathcal{Q}^{2}(F_b^{2})$. The key $\mathcal{K}^{1}$ and value $\mathcal{V}^{1}$ of $F_b^{1}$, and the key $\mathcal{K}^{2}$ and value $\mathcal{V}^{2}$ of $F_b^{2}$ are obtained using three generators $\mathcal{Q}$, $\mathcal{K}$, and $\mathcal{V}$. Then, $F_c^{1}\in\mathbb{R}^{L_a \times D_b \times H_b}$ obtained by cross-attention is defined as:
{
\begin{equation}
\resizebox{0.91\linewidth}{!}{$F_c^{1} = softmax(\mathcal{Q}^{1}(F_b^{1}) \otimes \mathcal{K}^{2}(F_b^{2})^T) \otimes \mathcal{V}^{2}(F_b^{2}),$}
\end{equation}
}
where the operation $T$ means matrix transpose, the operation $\otimes$ represents matrix multiplication, and the goal of $softmax$ function is to normalize the $F_c^{1}$. Finally, the cross-attention output feature $F_d^{1}\in\mathbb{R}^{L_a \times D_a}$ of the first branch is obtained by feature mapping. Similarly, we can get the cross-attention output $F_d^{2}\in\mathbb{R}^{L_a \times D_a}$ of the second branch. In this way, the model takes into account not only the meaning of each branch itself but also the relationships with other branch features, resulting in a more complete representation of multimodal data.

\begin{algorithm}[tb]
    \caption{Dynamic Selection Moudle Procedure}
    \label{alg_DSM}
    \textbf{Input}: Protein vector $X_{dsm}$ , Threshold $t$\\
    \textbf{Output}: Fusion feature after DSM
    \begin{algorithmic}[1] 
        \STATE Initialize expert weights $W \gets 0_N$. \\
        \STATE Compute expert confidence coefficients \\ \resizebox{0.45\linewidth}{!}{$\hat{p} \gets \text{Softmax}(\text{MLP}(X_{dsm}))$}. \\
        \STATE Select active experts $S \gets \{E_i | \hat{p}_i \geq t\}$.\\
        \FOR{each experts $E_i$ in $S$}{
         \STATE Normalize $\hat{p}$~to obtain weights $W_i \gets \frac{\hat{p}_i}{\sum_{E_j \in S} \hat{p}_j}$.
        }
        \ENDFOR
        \STATE \textbf{return} $\text{DSM}(X_{dsm}) \gets \text{Concat}(W_i \cdot E_i(X_{dsm}))$
    \end{algorithmic}
\end{algorithm}

\subsubsection{Dynamic Selection Module} 

In the final feature selection stage, we introduce DSM to enhance key features and mitigate the impact of conflicting ones. As illustrated in Algorithm \ref{alg_DSM} and Figure \ref{main}, this module employs an improved Mixture-of-Experts (MoE) strategy based on Masoudnia et al \cite{masoudnia2014mixture}. The MSL-Branch and MIL-Branch each output a single vector with three channels, where the three channels represent  PPI, sequence, and subcellular localization combined with domain features, respectively. All six-channel feature maps serve as the input $X_{dsm}=(x_{dsm}^1, x_{dsm}^2, \cdots, x_{dsm}^V)$ for the DSM. The function of DSM is:
\begin{equation}
\text{DSM}(X_{dsm})=\text{Concat}(\frac{\hat{p}_i}{\sum_{E_j \in S} \hat{p}_j} \cdot E_i(X_{dsm})),
\end{equation}

where $E_j$ is the experts belonging to the selected expert group S, $\hat{p}_i$ denotes the confidence coefficient of expert $E_i$.

\subsubsection{Loss Functions}
In this work, protein function prediction is modeled as the multi-label classification task. The predictor, constructed from fully connected layers, takes the output features of the DSM as input and produces an $M$-dimensional score vector of GO terms: $P_i=(p_{i}^1, p_{i}^2, \cdots, p_{i}^M))$.
In the context of protein function prediction using GO terms, there are significantly more negative proteins than positive ones in the training set. Consequently, we employ an asymmetric loss \cite{wu2023cfago} as the prediction loss $\mathcal{L}$. 
{
\begin{align}
\mathcal{L} =  & \frac{1}{NM} \sum_{i=1}^{N} \sum_{m=1}^{M} -y_i^m\left(1-p_i^m\right)^{y+} \log \left(p_i^m\right) \nonumber \\
& -\left(1-y_i^m\right)\left(p_i^m\right)^{y-} \log \left(1-p_i^m\right),
\end{align}
}
where $y_i^m$ represents the ground truth label for the $i$-th protein, while $p_i^m$ denotes the predicted score. The symbols $\{y+\}$ and $\{y-\}$ refer to the positive and negative focusing parameters respectively.

\section{Experiments}
In this section, we present the experimental setup, including the datasets, baseline models, training details, and evaluation metrics. Then we provide an analysis of the experimental results, supported by ablation studies and Davies-Bouldin scores to validate the effectiveness of the model.

Further experiments on the model components, structures, and parameters can be seen in Appendix Sections 1, 2, and 5.
\subsection{Experimental Setup}

\textbf{Dataset Settings.}
We construct our dataset based on CFAGO \cite{wu2023cfago}. PPI data comes from the STRING \cite{szklarczyk2023string} database (v11.5), and protein sequences, subcellular localization, and domain data are from the UniProt \cite{uniprot2023uniprot} database (v3.5.175). A total of 19,385 proteins are used for pretraining. For fine-tuning, we collect protein function annotations from the Gene Ontology \cite{gene2023gene} database (v2022-01-13). 
The fine-tuning datasets for each GO branch, split by two-time points, including BPO: 3,197 training, 304 validation, 182 testing proteins (45 GO terms), MFO: 2,747 training, 503 validation, 719 testing proteins (38 GO terms), and CCO: 5,263 training, 577 validation, 119 testing proteins (35 GO terms).

More details about sequence similarity and model performance are in Appendix Sections 3 and 6.

\textbf{Implementation Details.}
We conduct all experiments on NVIDIA GTX 4090. We set the dropout rate to 0.1 during pre-training, and the model trains for 5000 epochs, with a learning rate of 1e-5 for the first 2500 epochs and 1e-6 for the remaining 2500 epochs.
During fine-tuning, we use a dropout rate of 0.3 and train for 100 epochs with the AdamW optimizer. The learning rate is set to 1e-3 for the first 50 epochs and reduced to 1e-4 for the remaining 50 epochs.

\textbf{Compared Methods.} 
We compare DSRPGO with nine methods, which are categorized into two groups based on their data utilization strategies. Unimodal-based methods: Naive \cite{zhou2019cafa2}, BLAST\cite{altschul1990basic}, GeneMANIA\cite{mostafavi2008genemania}, Mashup\cite{cho2016compact}, and deepNF\cite{gligorijevic2018deepnf}. Multimodal-based methods: Graph2GO\cite{fan2020graph2go}, NetQuilt\cite{barot2021netquilt}, DeepGraphGO\cite{you2021deepgraphgo}, and CFAGO\cite{wu2023cfago}.

\textbf{Evaluation Metrics.}
In this study, we evaluate predictive performance using five metrics: micro-averaged AUPR (m-AUPR) and macro-averaged AUPR (M-AUPR) \cite{peng2021integrating}, F1-score (F1) \cite{wu2023cfago}, accuracy (ACC), and F-max score ($F_{\text{max}}$)\cite{bhzs}, providing a comprehensive assessment of model accuracy and effectiveness.

\begin{table*}
\centering
\fontsize{8pt}{8pt}\selectfont
{
\setlength{\tabcolsep}{1.0mm}
\renewcommand{\arraystretch}{0.9}
\begin{tabular}{cccccccccccc} 
\toprule
\multicolumn{2}{l}{Method}                & \multicolumn{1}{l}{$\text{Naïve}^\dag$} & \multicolumn{1}{l}{$\text{BLAST}^\dag$} & \multicolumn{1}{l}{$\text{GeneMANIA}^\dag$} & \multicolumn{1}{l}{$\text{Mashup}^\dag$} & \multicolumn{1}{l}{$\text{deepNF}^\dag$} & \multicolumn{1}{l}{NetQuilt} & \multicolumn{1}{l}{Graph2GO} & \multicolumn{1}{l}{DeepGraphGO} & \multicolumn{1}{l}{CFAGO} & \multicolumn{1}{l}{DSRPGO (Ours)}  \\ 
\midrule
\multirow{3}{*}{$\mathbf{F_{max}}$} & BPO & 0.051±0                   & 0.270±0                   & 0.000±0                           & 0.075±0                    & 0.394±0.006                & 0.164±0.014                  & 0.335±0.010                   & 0.327±0.028                     & \underline{0.439±0.007}       & \textbf{0.458±0.006}             \\ 
\cmidrule{2-2}
                                    & MFO & 0.177±0                   & 0.122±0                   & 0.000±0                           & 0.058±0                    & 0.153±0.004                & 0.081±0.013                  & 0.196±0.006                  & 0.142±0.035                     & \underline{0.236±0.004}       & \textbf{0.254±0.022}             \\ 
\cmidrule{2-2}
                                    & CCO & 0.121±0                   & 0.196±0                   & 0.031±0                       & 0.000±0                        & 0.297±0.009                & 0.138±0.013                  & 0.298±0.011                  & 0.209±0.023                     & \underline{0.366±0.018}       & \textbf{0.452±0.019}             \\ 
\cmidrule(lr){1-1}\cmidrule{2-2}
\multirow{3}{*}{m-AUPR}             & BPO & 0.024±0                   & 0.110±0                   & 0.042±0                       & 0.238±0                    & 0.303±0.006                & 0.077±0.006                  & 0.237±0.014                  & 0.210±0.022                     & \underline{0.328±0.005}       & \textbf{0.330±0.006}              \\ 
\cmidrule{2-2}
                                    & MFO & 0.050±0                   & 0.044±0                   & 0.050±0                       & 0.053±0                    & 0.089±0.001                & 0.045±0.007                  & 0.103±0.007                  & 0.080±0.021                     & \underline{0.159±0.003}       & \textbf{0.166±0.027}             \\ 
\cmidrule{2-2}
                                    & CCO & 0.047±0                   & 0.084±0                   & 0.103±0                       & 0.179±0                    & 0.178±0.005                & 0.081±0.003                  & 0.215±0.025                  & 0.133±0.011                     & \underline{0.337±0.005}       & \textbf{0.371±0.035}             \\ 
\cmidrule(lr){1-1}\cmidrule{2-2}
\multirow{3}{*}{M-AUPR}             & BPO & 0.048±0                   & 0.093±0                   & 0.160±0                       & 0.146±0                    & 0.174±0.005                & 0.081±0.004                  & 0.150±0.006                  & 0.133±0.008                     & \textbf{0.188±0.003}      & \underline{0.182±0.003}              \\ 
\cmidrule{2-2}
                                    & MFO & 0.029±0                   & 0.084±0                   & 0.109±0                       & 0.089±0                    & \underline{0.118±0.004}        & 0.064±0.003                  & 0.111±0.005                  & 0.098±0.007                     & \textbf{0.138±0.005}      & 0.114±0.009                      \\ 
\cmidrule{2-2}
                                    & CCO & 0.060±0                   & 0.082±0                   & 0.150±0                       & 0.104±0                    & 0.155±0.009                & 0.063±0.004                  & 0.159±0.021                  & 0.133±0.006                     & \underline{0.210±0.007}       & \textbf{0.239±0.025}             \\ 
\cmidrule(lr){1-1}\cmidrule{2-2}
\multirow{3}{*}{F1}                 & BPO & 0.035±0                   & 0.159±0                   & 0.054±0                       & 0.248±0                    & 0.228±0.005                & 0.114±0.017                  & 0.222±0.010                   & 0.238±0.012                     & \textbf{0.283±0.006}      & \underline{0.272±0.008}              \\ 
\cmidrule{2-2}
                                    & MFO & 0.004±0                   & 0.064±0                   & 0.008±0                       & 0.106±0                    & 0.117±0.004                & 0.070±0.016                  & 0.167±0.009                  & 0.165±0.056                     & \underline{0.234±0.005}       & \textbf{0.241±0.019}             \\ 
\cmidrule{2-2}
                                    & CCO & 0.070±0                   & 0.107±0                   & 0.123±0                       & 0.202±0                    & 0.205±0.009                & 0.108±0.013                  & 0.261±0.015                  & 0.210±0.016                     & \underline{0.314±0.007}       & \textbf{0.357±0.033}             \\ 
\cmidrule(lr){1-1}\cmidrule{2-2}
\multirow{3}{*}{ACC}                & BPO & 0.000±0                       & 0.071±0                   & 0.000±0                           & 0.044±0                    & 0.158±0.011                & 0.048±0.007                  & 0.257±0.007                  & 0.153±0.034                     & \underline{0.338±0.013}       & \textbf{0.346±0.016}             \\ 
\cmidrule{2-2}
                                    & MFO & 0.000±0                      & 0.015±0                   & 0.000±0                           & 0.038±0                    & 0.034±0.002                & 0.017±0.002                  & 0.114±0.015                  & 0.048±0.007                     & \underline{0.100±0.003}       & \textbf{0.124±0.037}             \\ 
\cmidrule{2-2}
                                    & CCO & 0.000±0                      & 0.034±0                   & 0.000±0                           & 0.000±0                        & 0.080±0.012                & 0.037±0.005                  & 0.180±0.024          & 0.066±0.011                     & \underline{0.210±0.008}       & \textbf{0.262±0.017}             \\

\bottomrule
\end{tabular}
}

\caption{Comparison results of different methods. Unimodal-based methods are marked with "$\dag$", while the rest are multimodal-based methods. The best results are highlighted in bold, and the sub-optimal results are underlined. After the ± is the standard deviation of the experimental results.}
\label{tab compare}
\end{table*}

\begin{figure}
    \centering
    \includegraphics[width=0.75\linewidth]{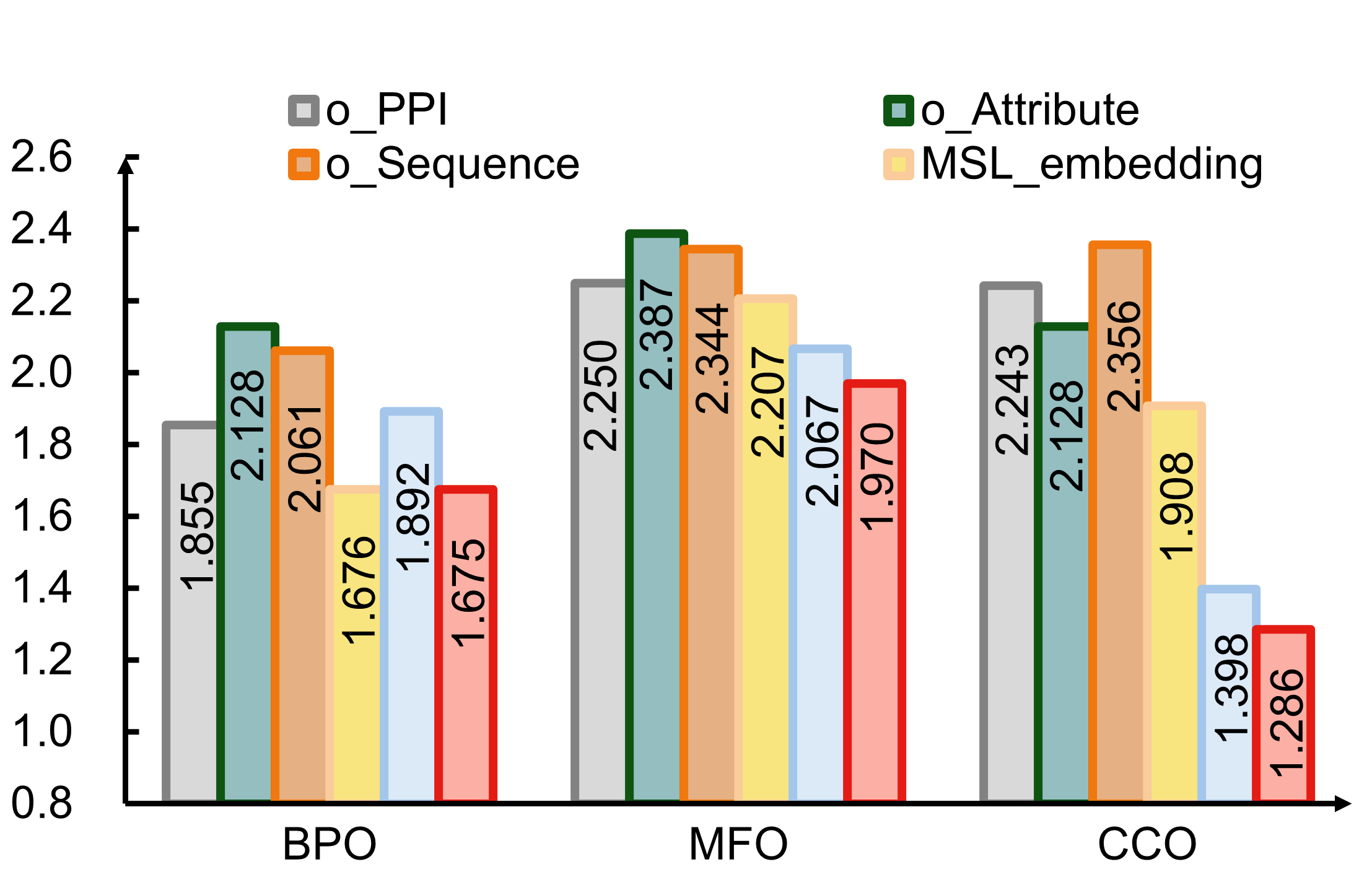}
    \caption{Davies Bouldin Score comparison of different protein features represents. o\_PPI, o\_Attribute, and o\_Sequence represent the original embedding of PPI, subcellular localization combined with domain, and protein language model, respectively. MSL\_embedding, MSI\_embedding, and DSM\_embedding represent the embedding from MSL-Branch, MIL-Branch, and DSM, respectively.}
    \label{db score}
\end{figure}

\begin{figure*}
    \centering
    \includegraphics[width=0.96\linewidth]{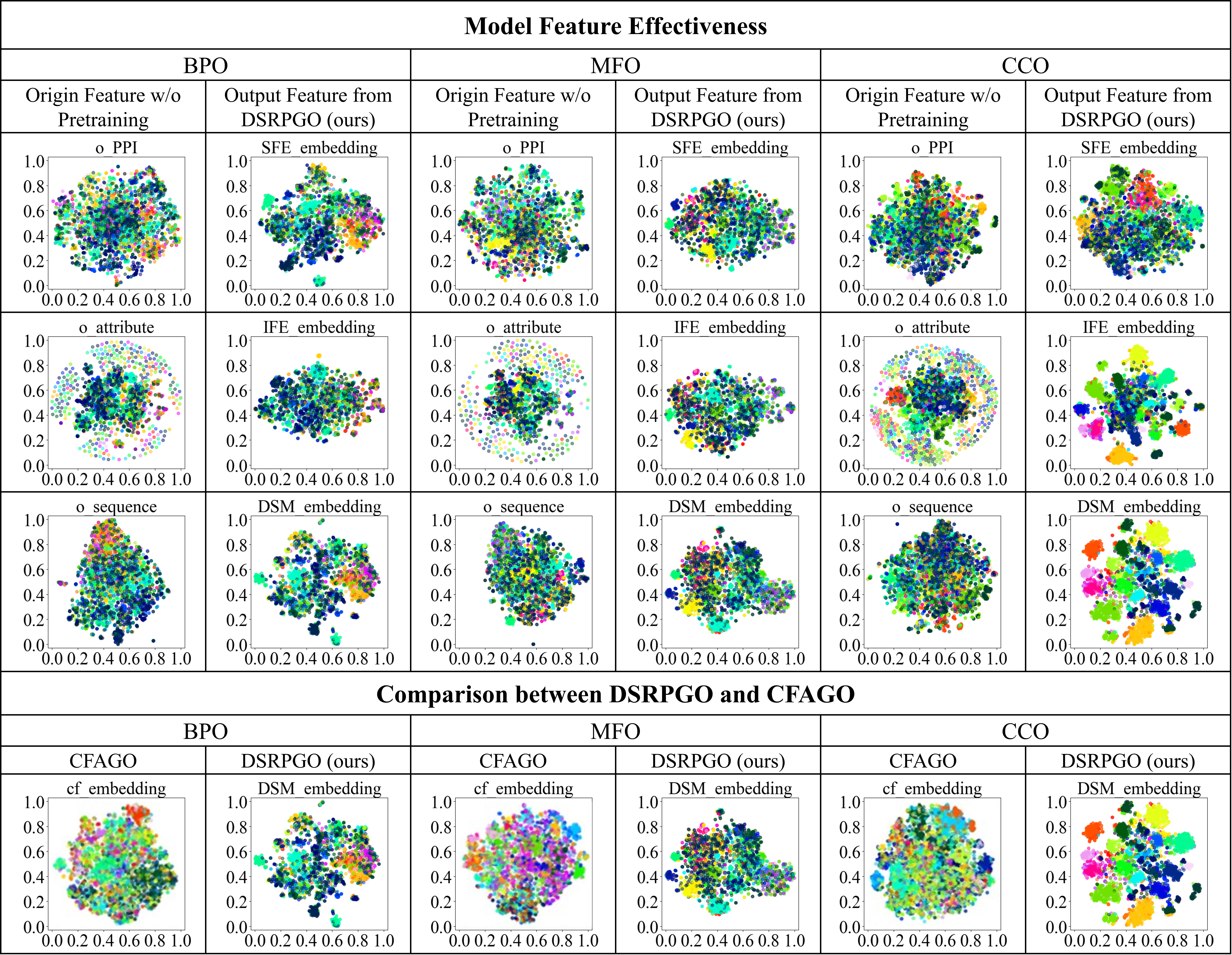}
    \caption{Visualization of different feature representations for DSRPGO, and comparison with CFAGO.}
    \label{tsn result}
\end{figure*}

\begin{table*}
\centering
\scriptsize
\setlength{\tabcolsep}{1mm}
\resizebox{\textwidth}{!}{
\begin{tabular}{l ccc ccc ccc ccc ccc}
\toprule
\multirow{2}{*}{\textbf{Method}} & \multicolumn{3}{c}{$\mathbf{F_{max}}$} & \multicolumn{3}{c}{\textbf{m-AUPR}}  & \multicolumn{3}{c}{\textbf{M-AUPR}} & \multicolumn{3}{c}{\textbf{F1}} & \multicolumn{3}{c}{\textbf{ACC}} \\
\cmidrule(lr){2-4} \cmidrule(lr){5-7} \cmidrule(lr){8-10} \cmidrule(lr){11-13} \cmidrule(lr){14-16}
& \textbf{BPO} & \textbf{MFO} & \textbf{CCO} & \textbf{BPO} & \textbf{MFO} & \textbf{CCO} & \textbf{BPO} & \textbf{MFO} & \textbf{CCO} & \textbf{BPO} & \textbf{MFO} & \textbf{CCO} & \textbf{BPO} & \textbf{MFO} & \textbf{CCO}\\
\midrule
MSLB                             & 0.437                     & 0.179                     & 0.371                     & 0.315                     & 0.108                     & 0.304                     & 0.173                     & 0.102                     & 0.197                     & 0.261                     & 0.172                     & 0.311                     & 0.292                     & 0.076                     & 0.190                      \\
MILB                             & 0.310                     & 0.179                     & 0.420                     & 0.180                     & 0.091                     & 0.330                     & 0.138                     & 0.113                     & 0.220                     & 0.236                     & 0.162                     & 0.342                     & 0.216                     & 0.090                     & 0.220                      \\
\textbf{MSLB+MILB}                        & \textbf{0.458}            & \textbf{0.254}            & \textbf{0.452}            & \textbf{0.330}            & \textbf{0.166}            & \textbf{0.371}            & \textbf{0.182}            & \textbf{0.114}            & \textbf{0.239}            & \textbf{0.272}            & \textbf{0.241}            & \textbf{0.357}            & \textbf{0.346}            & 0.124                     & \textbf{0.262}             \\
$w/o$ BInM                       & 0.435                     & 0.193                     & 0.333                     & 0.313                     & 0.116                     & 0.266                     & 0.174                     & 0.106                     & 0.186                     & 0.265                     & 0.180                     & 0.305                     & 0.301                     & 0.088                     & 0.151                      \\
$w/o$ DSM                        & 0.397                     & 0.190                     & 0.378                     & 0.275                     & 0.105                     & 0.302                     & 0.163                     & 0.113                     & 0.205                     & 0.265                     & 0.173                     & 0.328                     & 0.315                     & 0.092                     & 0.190                      \\
$w/o$ SP-F                       & 0.216                     & 0.173                     & 0.263                     & 0.106                     & 0.059                     & 0.164                     & 0.105                     & 0.039                     & 0.115                     & 0.174                     & 0.004                     & 0.226                     & 0.151                     & 0.000                     & 0.145                      \\
$w/o$ SE-F                       & 0.251                     & 0.238                     & 0.363                     & 0.119                     & 0.117                     & 0.219                     & 0.115                     & 0.099                     & 0.181                     & 0.179                     & 0.224                     & 0.322                     & 0.170                     & \textbf{0.133}            & 0.193                      \\
$w/o$ pretrain                         & \multicolumn{1}{l}{0.297} & \multicolumn{1}{l}{0.167} & \multicolumn{1}{l}{0.356} & \multicolumn{1}{l}{0.196} & \multicolumn{1}{l}{0.093} & \multicolumn{1}{l}{0.284} & \multicolumn{1}{l}{0.129} & \multicolumn{1}{l}{0.095} & \multicolumn{1}{l}{0.200} & \multicolumn{1}{l}{0.205} & \multicolumn{1}{l}{0.162} & \multicolumn{1}{l}{0.286} & \multicolumn{1}{l}{0.200} & \multicolumn{1}{l}{0.085} & \multicolumn{1}{l}{0.197}  \\
\bottomrule
\end{tabular}
}
\caption{Results of Ablation Studies. The overall model is denoted as 'MSLB+MILB’, where 'MSLB’ and 'MILB’ are the backbone components: MSL-Branch and MIL-Branch. $w/o$ BInM and $w/o$ DSM represent removing the BInM and DSM modules from the overall model. $w/o$ SP-F refers to removing spatial structure features from the input, while $w/o$ SE-F indicates removing sequence features. The best results are marked in bold.}
\label{ablation}
\end{table*}

\subsection{Comparison with Unimodal-based and Multimodal-based Methods}
\textbf{Comparision with Unimodal-based Methods.}
Most of the previous methods are based on unimodal protein features, so to verify the performance of our multimodal-based method, we compare our method with unimodal-based methods. The experimental results are shown in Table \ref{tab compare}. DSRPGO significantly outperforms unimodal-based methods across various metrics, except for M-AUPR in MFO. Compared to unimodal methods, DSRPGO improves Fmax by at least 6.4\% in BPO, 7.7\% in MFO, and 15.5\% in CCO. This demonstrates the advantage of integrating multimodal data for protein function prediction.

\textbf{Comparision with Multimodal-based Methods.}
To better evaluate our method, we also compare DSRPGO with other state-of-the-art multimodal-based methods, including CFAGO, DeepGraphGO, Graph2GO, and NetQuilt. The detailed results in Table \ref{tab compare} show that DSRPGO generally outperforms these methods. Compared to multimodal methods, DSRPGO improves the Fmax metric by at least 1.9\% in BPO, 1.8\% in MFO, and 8.6\% in CCO. This indicates that DSRPGO's architecture is more effective in learning deep representations among multimodal features, thereby further enhancing overall performance.
At the same time, we observe that DSRPGO does not perform optimally in M-AUPR. This is because M-AUPR evaluates each class equally, including those with fewer samples, which may not reflect the model’s overall performance. In contrast, m-AUPR aggregates performance across all classes, offering a more comprehensive measure of predictive capability.
In addition, we discuss the Structure-based and PLM-based comparison methods, as detailed in Appendix Section 4.

\subsection{Feature Effectiveness Analysis}
To further evaluate the distinguishing power of the multimodal features extracted by different components of DSRPGO, we use Davies-Bouldin (DB) \cite{wu2023cfago} scores. In the calculation of DB scores, GO terms are set as the labels for protein clusters, meaning proteins sharing the same GO term set are grouped into the same cluster. A lower DB score indicates more compact clusters and clearer separation. As shown in Figure \ref{db score}, DSRPGO components effectively capture multimodal features. Among them, DSM\_embedding performs best, indicating that DSM successfully integrates inputs from the MIL and MSL branches.

To further analyze the discriminative power of protein features, we visualize them using t-SNE \cite{chatzimparmpas2020t}, as shown in Figure \ref{tsn result}. Raw input features (o\_PPI, o\_Attribute, o\_Sequence), which are not pre-trained, show distinct patterns but lack clear clustering boundaries. In contrast, the output of the feature by various modules of DSRPGO achieves better clustering results. Additionally, compared to the output of the feature by CFAGO (cf\_embedding), DSRPGO demonstrates significantly superior performance.

\section{Ablation Studies}
In this section, the contributions of each component in DSRPGO are evaluated, as shown in Table \ref{ablation}.

\textbf{Analysis for Backbone Components.}
According to lines 1,2, and 3 of Table \ref{ablation}, the results of the backbone network only using MSL-Branch or MIL-Branch are not as good as those using combined branches. 

\textbf{Effectiveness of BInM.}
Considering the correlation of features among space and sequence, this method uses the BInM block to facilitate bidirectional multimodal feature interaction before DSM. As shown in rows 3 and 4 of Table \ref{ablation}, we verify the validity of BInM for the overall model by removing it.

\textbf{Effectiveness of DSM.}
To enable effective feature selection and accurate prediction of protein functions, DSM is used to select channel features most relevant to specific functional labels adaptively. At the same time, it reduces the interference and conflict caused by redundant features. As shown in rows 3 and 5 of Table \ref{ablation}, DSM has a positive impact on protein function prediction.

\textbf{Impact of Sequence and Spatial Structure Features.}
To verify the complementarity between sequence and spatial structure features, we perform an ablation study, retaining only spatial structure or sequence features. For the BInM module, it is removed as no interaction occurs with a single feature type. Rows 6 and 7 of Table \ref{ablation} show that removing feature interaction significantly reduces model performance.

\textbf{Impact of Pre-training.}
To evaluate the contribution of pre-training, we conduct an ablation study by removing it. As shown in the last row of Table \ref{ablation}, the model's performance drops significantly across all metrics without pre-training.

\section{Conclusion}

This paper proposes a dual-branched multimodal method for protein function prediction with reconstructive pre-training. The proposed method enhances the model's ability to integrate multimodal features through two key components: the BInM and the DSM, leading to significant performance gains. Experimental results show that the DSRPGO outperforms current state-of-the-art unimodal and multimodal methods across multiple metrics. These results underscore the importance of integrating multimodal data to enhance protein function prediction, and validate the superiority of the BInM and the DSM in multimodal protein data integration. 

\section*{Acknowledgements}
This work was supported in part by the National Natural Science Foundation of China under Grant
No. 62302317, the Natural Science Foundation of Guangdong Province under Grant 2025A1515010184, the project of Shenzhen Science and Technology Innovation Committee under Grant JCYJ20240813141424032 and JCYJ20240813112420027, and the Foundation for Young innovative talents in ordinary universities of Guangdong under Grant 2024KQNCX042, the Stable Support Projects for Shenzhen Higher Education Institutions under grant 20231122005530001 and 20220715183602001, and Guangdong Basic and Applied Basic Research Foundation grant 2024A1515220079.

\section*{Contribution Statement}
Xiaoling Luo and Peng Chen contributed equally to this work. 
\bibliographystyle{named}
\bibliography{ijcai25}

\begin{thebibliography}{}

\bibitem[\protect\citeauthoryear{Aleksander \bgroup \em et al.\egroup }{2023}]{gene2023gene}
Suzi~A Aleksander, James Balhoff, Seth Carbon, J~Michael Cherry, Harold~J Drabkin, Dustin Ebert, Marc Feuermann, Pascale Gaudet, Nomi~L Harris, et~al.
\newblock The gene ontology knowledgebase in 2023.
\newblock {\em Genetics}, 224(1):iyad031, 2023.

\bibitem[\protect\citeauthoryear{Altschul \bgroup \em et al.\egroup }{1990}]{altschul1990basic}
Stephen~F Altschul, Warren Gish, Webb Miller, Eugene~W Myers, and David~J Lipman.
\newblock Basic local alignment search tool.
\newblock {\em Journal of Molecular Biology}, 215(3):403--410, 1990.

\bibitem[\protect\citeauthoryear{Barot \bgroup \em et al.\egroup }{2021}]{barot2021netquilt}
Meet Barot, Vladimir Gligorijevi{\'c}, Kyunghyun Cho, and Richard Bonneau.
\newblock Netquilt: deep multispecies network-based protein function prediction using homology-informed network similarity.
\newblock {\em Bioinformatics}, 37(16):2414--2422, 2021.

\bibitem[\protect\citeauthoryear{Chatzimparmpas \bgroup \em et al.\egroup }{2020}]{chatzimparmpas2020t}
Angelos Chatzimparmpas, Rafael~M Martins, and Andreas Kerren.
\newblock t-visne: Interactive assessment and interpretation of t-sne projections.
\newblock {\em IEEE Transactions on Visualization and Computer Graphics}, 26(8):2696--2714, 2020.

\bibitem[\protect\citeauthoryear{Chen and Luo}{2024}]{chen2024dualnetgo}
Zhuoyang Chen and Qiong Luo.
\newblock Dualnetgo: a dual network model for protein function prediction via effective feature selection.
\newblock {\em Bioinformatics}, 40(7), 2024.

\bibitem[\protect\citeauthoryear{Cho \bgroup \em et al.\egroup }{2016}]{cho2016compact}
Hyunghoon Cho, Bonnie Berger, and Jian Peng.
\newblock Compact integration of multi-network topology for functional analysis of genes.
\newblock {\em Cell Systems}, 3(6):540--548, 2016.

\bibitem[\protect\citeauthoryear{Consortium}{2022}]{uniprot2023uniprot}
The~UniProt Consortium.
\newblock {UniProt: the Universal Protein Knowledgebase in 2023}.
\newblock {\em Nucleic Acids Research}, 51(D1):D523--D531, 2022.

\bibitem[\protect\citeauthoryear{Dosovitskiy \bgroup \em et al.\egroup }{2021}]{Dosovitskiy2021}
A.~Dosovitskiy, L.~Beyer, A.~Kolesnikov, D.~Weissenborn, X.~Zhai, T.~Unterthiner, M.~Dehghani, M.~Minderer, G.~Heigold, S~Gelly, J.~Uszkoreit, and N~Houlsby.
\newblock An image is worth 16x16 words: Transformers for image recognition at scale.
\newblock In {\em International Conference on Learning Representations}, 2021.

\bibitem[\protect\citeauthoryear{Elnaggar \bgroup \em et al.\egroup }{2021}]{prott5}
Ahmed Elnaggar, Michael Heinzinger, Christian Dallago, Ghalia Rehawi, Wang Yu, Llion Jones, Tom Gibbs, Tamas Feher, Christoph Angerer, Martin Steinegger, Debsindhu Bhowmik, and Burkhard Rost.
\newblock Prottrans: Towards cracking the language of lifes code through self-supervised deep learning and high performance computing.
\newblock {\em IEEE Transactions on Pattern Analysis and Machine Intelligence}, pages 1--1, 2021.

\bibitem[\protect\citeauthoryear{Fan \bgroup \em et al.\egroup }{2020}]{fan2020graph2go}
Kunjie Fan, Yuanfang Guan, and Yan Zhang.
\newblock Graph2go: a multi-modal attributed network embedding method for inferring protein functions.
\newblock {\em GigaScience}, 9(8):giaa081, 2020.

\bibitem[\protect\citeauthoryear{Gligorijevi{\'c} \bgroup \em et al.\egroup }{2018}]{gligorijevic2018deepnf}
Vladimir Gligorijevi{\'c}, Meet Barot, and Richard Bonneau.
\newblock deepnf: deep network fusion for protein function prediction.
\newblock {\em Bioinformatics}, 34(22):3873--3881, 2018.

\bibitem[\protect\citeauthoryear{Gu and Dao}{2023}]{gu2023mamba}
Albert Gu and Tri Dao.
\newblock Mamba: Linear-time sequence modeling with selective state spaces.
\newblock {\em arXiv preprint arXiv:2312.00752}, 2023.

\bibitem[\protect\citeauthoryear{Gu \bgroup \em et al.\egroup }{2023}]{gu2022train}
Albert Gu, Isys Johnson, Aman Timalsina, Atri Rudra, and Christopher R{\'e}.
\newblock How to train your hippo: State space models with generalized orthogonal basis projections.
\newblock In {\em Proceedings of the International Conference on Learning Representations}, 2023.

\bibitem[\protect\citeauthoryear{Hasselgren and Oprea}{2024}]{hasselgren2024artificial}
Catrin Hasselgren and Tudor~I Oprea.
\newblock Artificial intelligence for drug discovery: Are we there yet?
\newblock {\em Annual Review of Pharmacology and Toxicology}, 64(1):527--550, 2024.

\bibitem[\protect\citeauthoryear{Kulmanov and Hoehndorf}{2020}]{kulmanov2020deepgoplus}
Maxat Kulmanov and Robert Hoehndorf.
\newblock Deepgoplus: improved protein function prediction from sequence.
\newblock {\em Bioinformatics}, 36(2):422--429, 2020.

\bibitem[\protect\citeauthoryear{Lin \bgroup \em et al.\egroup }{2024}]{bhzs}
Baohui Lin, Xiaoling Luo, Yumeng Liu, and Xiaopeng Jin.
\newblock {A comprehensive review and comparison of existing computational methods for protein function prediction}.
\newblock {\em Briefings in Bioinformatics}, 25(4):bbae289, 2024.

\bibitem[\protect\citeauthoryear{Ma \bgroup \em et al.\egroup }{2025}]{MA2025125366}
Ruixin Ma, Longfei Wang, Huinan Wu, Buyun Gao, Xiaoru Wang, and Liang Zhao.
\newblock Historical trends and normalizing flow for one-shot temporal knowledge graph reasoning.
\newblock {\em Expert Systems with Applications}, 260:125366, 2025.

\bibitem[\protect\citeauthoryear{Masoudnia and Ebrahimpour}{2014}]{masoudnia2014mixture}
Saeed Masoudnia and Reza Ebrahimpour.
\newblock Mixture of experts: a literature survey.
\newblock {\em Artificial Intelligence Review}, 42:275--293, 2014.

\bibitem[\protect\citeauthoryear{Mostafavi \bgroup \em et al.\egroup }{2008}]{mostafavi2008genemania}
Sara Mostafavi, Debajyoti Ray, David Warde-Farley, Chris Grouios, and Quaid Morris.
\newblock Genemania: a real-time multiple association network integration algorithm for predicting gene function.
\newblock {\em Genome Biology}, 9:1--15, 2008.

\bibitem[\protect\citeauthoryear{Pan \bgroup \em et al.\egroup }{2023}]{pan2023pfresgo}
Tong Pan, Chen Li, Yue Bi, Zhikang Wang, Robin~B Gasser, Anthony~W Purcell, Tatsuya Akutsu, Geoffrey~I Webb, Seiya Imoto, and Jiangning Song.
\newblock Pfresgo: an attention mechanism-based deep-learning approach for protein annotation by integrating gene ontology inter-relationships.
\newblock {\em Bioinformatics}, 39(3):btad094, 2023.

\bibitem[\protect\citeauthoryear{Paysan-Lafosse \bgroup \em et al.\egroup }{2023}]{paysan2023interpro}
Typhaine Paysan-Lafosse, Matthias Blum, Sara Chuguransky, Tiago Grego, Beatriz~L{\'a}zaro Pinto, Gustavo~A Salazar, Maxwell~L Bileschi, Peer Bork, Alan Bridge, Lucy Colwell, et~al.
\newblock Interpro in 2022.
\newblock {\em Nucleic Acids Research}, 51(D1):D418--D427, 2023.

\bibitem[\protect\citeauthoryear{Peng \bgroup \em et al.\egroup }{2021}]{peng2021integrating}
Jiajie Peng, Hansheng Xue, Zhongyu Wei, Idil Tuncali, Jianye Hao, and Xuequn Shang.
\newblock Integrating multi-network topology for gene function prediction using deep neural networks.
\newblock {\em Briefings in bioinformatics}, 22(2):2096--2105, 2021.

\bibitem[\protect\citeauthoryear{Radivojac \bgroup \em et al.\egroup }{2013}]{zhou2019cafa2}
Predrag Radivojac, Wyatt~T Clark, Tal~Ronnen Oron, Alexandra~M Schnoes, Tobias Wittkop, Artem Sokolov, Kiley Graim, Christopher Funk, Karin Verspoor, Asa Ben-Hur, et~al.
\newblock A large-scale evaluation of computational protein function prediction.
\newblock {\em Nature Methods}, 10(3):221--227, 2013.

\bibitem[\protect\citeauthoryear{Szklarczyk \bgroup \em et al.\egroup }{2023}]{szklarczyk2023string}
Damian Szklarczyk, Rebecca Kirsch, Mikaela Koutrouli, Katerina Nastou, Farrokh Mehryary, Radja Hachilif, Annika~L Gable, Tao Fang, Nadezhda~T Doncheva, Sampo Pyysalo, et~al.
\newblock The string database in 2023: protein--protein association networks and functional enrichment analyses for any sequenced genome of interest.
\newblock {\em Nucleic Acids Research}, 51(D1):D638--D646, 2023.

\bibitem[\protect\citeauthoryear{Wang \bgroup \em et al.\egroup }{2022}]{9387128}
Jun Wang, Long Zhang, An~Zeng, Dawen Xia, Jiantao Yu, and Guoxian Yu.
\newblock Deepiii: Predicting isoform-isoform interactions by deep neural networks and data fusion.
\newblock {\em IEEE/ACM Transactions on Computational Biology and Bioinformatics}, 19(4):2177--2187, 2022.

\bibitem[\protect\citeauthoryear{Wu \bgroup \em et al.\egroup }{2023}]{wu2023cfago}
Zhourun Wu, Mingyue Guo, Xiaopeng Jin, Junjie Chen, and Bin Liu.
\newblock Cfago: cross-fusion of network and attributes based on attention mechanism for protein function prediction.
\newblock {\em Bioinformatics}, 39(3):btad123, 2023.

\bibitem[\protect\citeauthoryear{You \bgroup \em et al.\egroup }{2021}]{you2021deepgraphgo}
Ronghui You, Shuwei Yao, Hiroshi Mamitsuka, and Shanfeng Zhu.
\newblock Deepgraphgo: graph neural network for large-scale, multispecies protein function prediction.
\newblock {\em Bioinformatics}, 37(Supplement\_1):i262--i271, 2021.

\bibitem[\protect\citeauthoryear{Zhang \bgroup \em et al.\egroup }{2023}]{zhang2023hnetgo}
Xiaoshuai Zhang, Huannan Guo, Fan Zhang, Xuan Wang, Kaitao Wu, Shizheng Qiu, Bo~Liu, Yadong Wang, Yang Hu, and Junyi Li.
\newblock Hnetgo: protein function prediction via heterogeneous network transformer.
\newblock {\em Briefings in Bioinformatics}, 24(6):bbab556, 2023.

\bibitem[\protect\citeauthoryear{Zhao \bgroup \em et al.\egroup }{2024}]{10822064}
Liang Zhao, Jian Zhang, Bo~Xu, Yi~Yang, Yangqianhui Zhang, and Ruixin Ma.
\newblock Multimodal contrastive learning with neuroimaging and cognitive tests for alzheimer’s disease diagnosis.
\newblock In {\em 2024 IEEE International Conference on Bioinformatics and Biomedicine (BIBM)}, volume~1, pages 2971--2976, 2024.

\bibitem[\protect\citeauthoryear{Zhou \bgroup \em et al.\egroup }{2019}]{8983075}
Guangjie Zhou, Jun Wang, Xiangliang Zhang, and Guoxian Yu.
\newblock Deepgoa: Predicting gene ontology annotations of proteins via graph convolutional network.
\newblock In {\em Proceedings of the IEEE International Conference on Bioinformatics and Biomedicine (BIBM)}, volume~1, pages 1836--1841, 2019.

\bibitem[\protect\citeauthoryear{Zhu \bgroup \em et al.\egroup }{2024}]{zhu2024vision}
Lianghui Zhu, Bencheng Liao, Qian Zhang, Xinlong Wang, Wenyu Liu, and Xinggang Wang.
\newblock Vision mamba: efficient visual representation learning with bidirectional state space model.
\newblock In {\em Proceedings of the 41st International Conference on Machine Learning}, 2024.

\end{thebibliography}

\end{document}